\documentclass[conference]{IEEEtran}
\IEEEoverridecommandlockouts
\usepackage{cite}
\usepackage{amsmath,amssymb,amsfonts}
\usepackage{algorithmic}
\usepackage{graphicx}
\usepackage{textcomp}
\usepackage{xcolor}
\usepackage{url}
\usepackage{tcolorbox}
\tcbuselibrary{raster,skins,breakable}
\tcbuselibrary{xparse} 
\DeclareTColorBox{brekableitembox}{ o m O{.5} O{} }%
  {empty, left=2mm, right=2mm, top=-2mm, bottom=0.2mm, attach boxed title to top left={xshift=1.2em},
  boxed title style={empty,left=-2mm,right=-2mm}, colframe=black, coltitle=black, coltext=black, breakable,  
  underlay unbroken={\draw[black,line width=#3pt]
    (title.east) -- (title.east-|frame.east) -- (frame.south east) -- (frame.south west) -- (title.west-|frame.west) -- (title.west); },
  underlay first={\draw[black,line width=#3pt](title.east) -- (title.east-|frame.east) -- (frame.south east) ;
    \draw[black,line width=#3pt] (frame.south west) -- (title.west-|frame.west) -- (title.west); },
  underlay middle={\draw[black,line width=#3pt](frame.north east) -- (frame.south east) ;
    \draw[black,line width=#3pt](frame.south west) -- (frame.north west) ;},
  underlay last={\draw[black,line width=#3pt](frame.north east) -- (frame.south east) -- (frame.south west) -- (frame.north west) ;}, IfValueTF={#1}{title=~~#2~~〈#1〉}{title=~~#2~~},#4}

\def\BibTeX{{\rm B\kern-.05em{\sc i\kern-.025em b}\kern-.08em
    T\kern-.1667em\lower.7ex\hbox{E}\kern-.125emX}}
\begin{document}
\bstctlcite{IEEEexample:BSTcontrol}

\title{Construction of Domain-specified\\Japanese Large Language Model for Finance\\through Continual Pre-training}

\author{
    \IEEEauthorblockN{Masanori Hirano}
    \IEEEauthorblockA{
        \textit{Preferred Networks, Inc.}\\
        Tokyo, Japan \\
        research@mhirano.jp}
    \and
    \IEEEauthorblockN{Kentaro Imajo}
    \IEEEauthorblockA{
        \textit{Preferred Networks, Inc.}\\
        Tokyo, Japan \\
        imos@preferred.jp}

}

\maketitle

\begin{abstract}
    Large language models (LLMs) are now widely used in various fields, including finance.
    However, Japanese financial-specific LLMs have not been proposed yet.
    Hence, this study aims to construct a Japanese financial-specific LLM through continual pre-training.
    Before tuning, we constructed Japanese financial-focused datasets for continual pre-training.
    As a base model, we employed a Japanese LLM that achieved state-of-the-art performance on Japanese financial benchmarks among the 10-billion-class parameter models.
    After continual pre-training using the datasets and the base model, the tuned model performed better than the original model on the Japanese financial benchmarks.
    Moreover, the outputs comparison results reveal that the tuned model's outputs tend to be better than the original model's outputs in terms of the quality and length of the answers.
    These findings indicate that domain-specific continual pre-training is also effective for LLMs.
    The tuned model is publicly available on Hugging Face.
\end{abstract}


\section{Introduction}
Recently, large language models (LLMs) have demonstrated excellent performance.
In particular, the latest models, such as ChatGPT\cite{chatgpt} and GPT-4\cite{GPT4}, exhibit high performance and significant generalization abilities.
The basis of these models begins with the transformer \cite{Vaswani2017} and BERT\cite{Devlin2018}, and GPT series \cite{GPT-1,GPT-2,GPT-3} were developed using the transformer.
Other LLMs have also been proposed, such as Bard\cite{bard}, LLaMA\cite{touvron2023llama,Touvron2023}, Dolly\cite{dolly}, BLOOM\cite{scao2022bloom}, Vicuna\cite{vicuna}, PaLM\cite{Chowdhery2022,Anil2023}, and Gemini \cite{gemini}.

The major difference between the latest LLMs and previous language models (e.g., BERT) is that one model can answer questions in multiple languages and domains and respond by following the instructions.
Previously, BERT was trained separately in different languages and domains \cite{Suzuku2023-ipm}.
However, the latest LLMs, such as GPT4, can freely process multiple languages.
Moreover, whereas BERT can only fill in incomplete sentences, the latest LLMs can answer questions in the same manner as humans.

Even if LLM can answer questions in multiple languages and domains, domain-specific models could still be useful.
For example, Hirano {\it et al.} \cite{Hirano2023-nbis} tuned the English-based model to Japanese and achieved better outputs than the original model.
Sukeda {\it et al.} \cite{sukeda2023jmedlora} also tuned the English-based model to the Japanese medical domain.
Back to the era of BERT, SciBERT \cite{beltagy2019scibert}, MedBERT \cite{rasmy2021med}, Japanese BERT\footnote{\url{https://huggingface.co/ ohoku-nlp/bert-base-japanese}}, and Japanese financial BERT \cite{Suzuki2022-sigfin28} are proposed.
Moreover, Howard {\it et al.} \cite{howard2018ulmfit} proposed universal language model fine-tuning and the methodologies, and effects of domain-specified fine-tuning were discussed in\cite{gururangan2020don,Suzuku2023-ipm}.

In this study, we try to construct a Japanese financial-specific LLM.
Financial services are now hot topics in the use of LLMs.
For instance, BloombergGPT\cite{Wu2023} is a private LLM focused on finance.
In addition, publicly available models, such as FinLLAMA\cite{Fin-LLAMA}, which is a tuned version of LLaMA\cite{touvron2023llama}, FinGPT\cite{yang2023fingpt}, and Instruct-FinGPT\cite{zhang2023instruct}, exist.
However, Japanese financial-specific LLMs have yet to be proposed.
Moreover, Japanese-focused LLM benchmarks have already been constructed \cite{Hirano2023-finnlpkdf}.
Therefore, it is high time that a Japanese financial-specific LLM is constructed.

This study employs a domain-specific (financial-specific) continual pre-training on an existing Japanese LLM and checks if the model performance on the Japanese financial benchmarks \cite{Hirano2023-finnlpkdf} improves or not.
The existing Japanese LLM we employed in this study is rinna/nekomata-14b, which is publicly available on Hugging Face\footnote{\url{https://huggingface.co/rinna/nekomata-14b}} and achieved the state-of-the-art performance on Japanese financial benchmarks among the 10-billions-class-parameters models (13b/14b models).

Consequently, the tuned model performed better than the original model on the Japanese financial benchmarks.
This means that the domain-specific continual pre-training is effective for the Japanese financial-specific LLM.

The tuned model is publicly available on Hugging Face: \url{https://huggingface.co/pfnet/nekomata-14b-pfn-qfin}.

\section{Related Work}
Studies on specialized language models in finance and Japanese have been conducted for a long time.
The classic vector embedding technique used in language processing is word2vec \cite{Mikolov2013a}.
Word2vec has also been used in the financial domain \cite{Hirano2019-information}.
After word2vec, ELMo \cite{peters2018elmo}, which uses a bidirectional long short-term memory (LSTM) \cite{schuster1997bilstm} to pre-train a distributed representation, appeared, along with transformer \cite{Vaswani2017}, which is a good alternative to LSTM in time-series processing, and transformer-based BERT \cite{Devlin2018}.

In contrast, methodologies to fit language models to specific languages or domains are also pursued.
For instance, Howard {\it et al.} \cite{howard2018ulmfit} proposed universal language model fine-tuning.
Following this study, some domain- or language-specific language models were developed, such as SciBERT \cite{beltagy2019scibert}, MedBERT \cite{rasmy2021med}, Japanese BERT\footnote{\url{https://huggingface.co/tohoku-nlp/bert-base-japanese}}, and Japanese financial BERT \cite{Suzuki2022-sigfin28}.
Moreover, the methodologies and effects of domain-specified fine-tuning were discussed in\cite{gururangan2020don,Suzuku2023-ipm}.

In the era of LLMs, although several transformer-based language models have been proposed, as described in the Introduction section, several unknown mechanisms of LLMs exist, and numerous trials have been performed.

Several proposed LLMs that focus specifically on finance exist.
For instance, BloombergGPT\cite{Wu2023} is a private LLM focused on finance.
In addition, publicly available models, such as FinLLAMA\cite{Fin-LLAMA}, which is a tuned version of LLaMA\cite{touvron2023llama}, FinGPT\cite{yang2023fingpt}, and Instruct-FinGPT\cite{zhang2023instruct}, exist.

Japanese-focused LLMs and benchmarks have also been developed.
Various models such as CyberAgent's CALM series, Rinna's model, stabilityai's stablelm series, Elyza's model, Preferred Networks' Plamo\texttrademark, and LLM-jp-13B have been proposed.
However, few models have been published in academic research papers.
Other studies have tuned existing English-based models to specialize in Japanese-language use\cite{Hirano2023-nbis,sukeda2023jmedlora,suzuki2023base}.
As for the Japanese task evaluation for LLMs, several benchmarks are available, including the jlm\_eval\cite{jlm_eval}, llm-jp-eval\cite{llm-jp-eval}, and Rakuda benchmarks\footnote{\url{https://yuzuai.jp/benchmark}}.
Moreover, the Japanese financial benchmarks have been constructed \cite{Hirano2023-finnlpkdf}.

Some possible tuning methods for LLMs are available.
For instance, Low-Rank adaptation \cite{hu2021lora} could be one possible method for domain-specific tuning.
Moreover, other tuning methods, such as instruction tuning \cite{wei2021finetuned}, reinforcement learning from human preferences \cite{christiano2017deep}, and direct preference optimization \cite{rafailov2024direct} are also proposed.
However, according to superficial alignment hypothesis \cite{zhou2024lima}, those tuning methods might not be effective for domain-specific tuning because the tuning focusing on the alignment cannot learn new knowledge.
Therefore, we employed the continual pre-training method for domain-specific tuning in this study.

\section{Tuning Method and Experiments}
In this study, we employed the continual pre-training method for domain-specific tuning.
To run an experiment, we need an existing Japanese LLM, Japanese financial datasets, and a Japanese financial benchmark.
We describe their details in the following subsections.

\subsection{Continual Pre-training for Domain-specific Tuning}
As a tuning method, we employed the continual pre-training method.
This is because the continual pre-training method is effective for domain-specific tuning, as described in the Related Work section.

As a base model, we employed rinna/nekomata-14b, publicly available on Hugging Face\footnote{\url{https://huggingface.co/rinna/nekomata-14b}}.
The rinna/nekomata-14b model is a Japanese LLM that achieved state-of-the-art performance on Japanese financial benchmarks among the 10-billions-class-parameters models (13b/14b models).
If we want to reveal the effectiveness of the domain-specific tuning, we need to employ the state-of-the-art model as a base model.

For the tuning, we employed the accelerate library\cite{accelerate} with deepspeed \cite{rasley2020deepspeed} to enable data-parallelized distributed training.
The other hyperparameters were set as the following:
\begin{itemize}
    \item Devices: A100 80GB x4
    \item Learning rate: starting from 5e-7, and decayed linearly to 0
    \item Number of epochs: 5
    \item Batch size: 24 (6 per device)
    \item Max sequence length: 2048
    \item Dtype: bf16
    \item Gradient accumulation steps: 1
    \item Gradient checkpointing: True
\end{itemize}

\subsection{Japanese Financial Focused Datasets}
To tune the model, we constructed Japanese financial-focused datasets for pre-training.
Different from instruction tuning \cite{wei2021finetuned}, we employed the continual pre-training method.
Therefore, different from the instruction dataset, the datasets should contain various raw financial documents.

For the datasets, we crawled some articles from the Internet and cleaned them.
The datasets are currently clear to use for commercial purposes under Japanese law as of April 2024.
The crawled articles mainly include the following types of documents:
\begin{itemize}
    \item Speeches, Press Conferences, and Talks of Officers of the Bank of Japan
    \item Minutes of the Monetary Policy Meetings of the Bank of Japan
    \item Reports, glossaries, and company profiles from multiple financial institutions
    \item Financial-related documents extracted from Wikipedia (using Wikipedia dumps)
\end{itemize}

Moreover, the following official published documents were also included via their API services:
\begin{itemize}
    \item Reports on EDInet\footnote{\url{https://disclosure2.edinet-fsa.go.jp/}}
\end{itemize}

Those documents were cleansed and formatted mainly in the following formats:
\begin{itemize}
    \item Plain markdown format (converted from HTML/PDF)
    \item Section-wise consolidated format
    \item Category-wise consolidated format (including category/keyword name, description, and corresponding stocks)
    \item List format (Company name, its stock code, and its industry in each line)
    \item Question-and-answer format (One question and its answer)
    \item Multiple choice question format (one question, its multiple choices, and the correct answer)
\end{itemize}
For the formatting, stabilityai's japanese-stablelm-base-gamma-7b\footnote{\url{https://huggingface.co/stabilityai/japanese-stablelm-base-gamma-7b}} is partly used.
Especially the question-and-answer format and the multiple choice questions are generated with almost the same approach as WRAP \cite{maini2024rephrasing}.
The final datasets contain about 8.1 million documents and 370 million tokens.

\subsection{Financial Focused Evaluation}
We employed two types of evaluation methods to evaluate the tuned model.
\begin{itemize}
    \item Benchmark evaluation: We employed the Japanese financial benchmarks \cite{Hirano2023-finnlpkdf} for evaluating the model. This is a quantitative evaluation.
    \item Outputs comparison: We compared the outputs of the tuned model with the original model. This is a qualitative evaluation.
\end{itemize}

The Japanese financial benchmarks \cite{Hirano2023-finnlpkdf} is currently the most popular benchmark for evaluating the Japanese LLMs in financial services.
The benchmark contains the following tasks:
\begin{itemize}
    \item chabsa: Aspect-based sentiment analysis
    \item cma\_basics: Fundamental knowledge questions in securities analysis
    \item cpa\_audit: Japanese Certified Public Accountant (CPA) exam, which comes from \cite{Masuda2023}
    \item fp2: 2nd grade Japanese financial planner exam
    \item security\_sales\_1: 1st-grade Japanese securities broker representative test
\end{itemize}
Almost all tasks are multiple-choice questions, and the answers are evaluated by the F1 score (for Chabsa) or accuracy (for others).
In benchmark evaluation, we employed the following settings:
\begin{itemize}
    \item Prompts: Default prompts of the benchmarks (chabsa, cma\_basics, cpa\_audit, fp2, security\_sales\_1)
    \item \# of fewshots: 0
\end{itemize}
Those settings are employed for simplification and fair comparison with the original model.

In the outputs comparison, we generated the outputs of the tuned model and the original model for the same prompts.
Subsequently, we compared the outputs and checked whether the tuned model's outputs were better than the original model's outputs in terms of the quality of the answers.

In the outputs comparison, we employed the following settings:
\begin{itemize}
    \item Max new tokens: 512
    \item Sampling: False
    \item Top-k: 50
    \item Repetition penalty: 1.1
\end{itemize}

However, the output comparison is a subjective evaluation.
Therefore, we employed the benchmark evaluation for the quantitative evaluation, and the outputs comparison is mainly aimed at making it easier to understand the effectiveness of the tuning for the readers.

\section{Results}

\begin{figure}[htb]
    \centering
    \includegraphics[width=\linewidth]{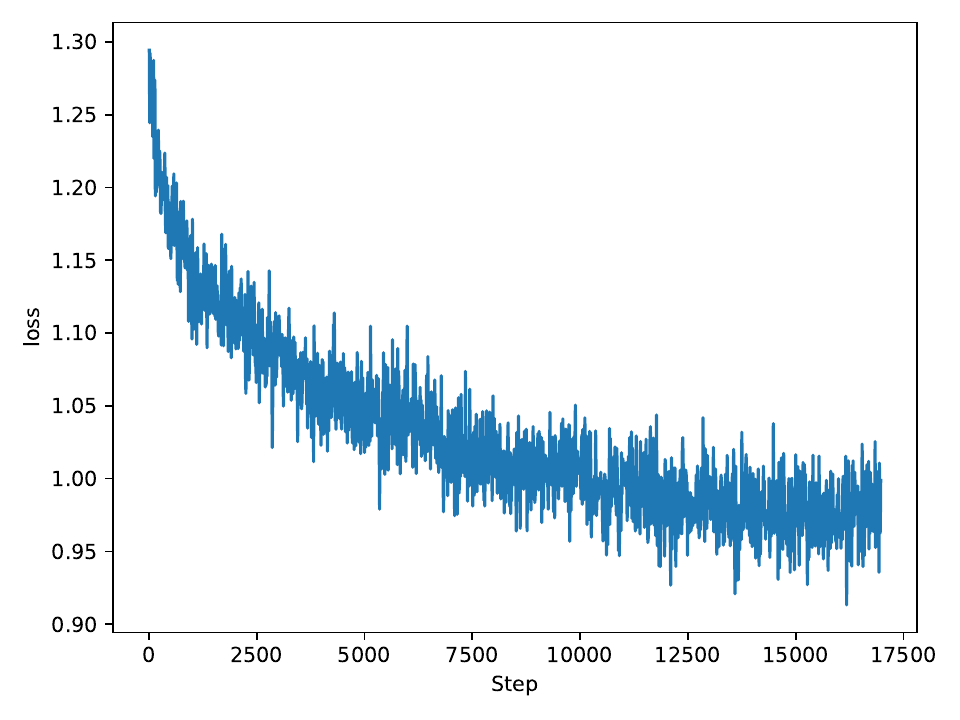}
    \caption{Loss Curve}
    \label{fig:loss}
\end{figure}
In figure \ref{fig:loss}, the loss curve of the continual pre-training is shown.
In our tuning, no loss spikes were observed.
The loss curve was also saturated as the learning rate decayed linearly to 0.

In the following, we show the benchmark evaluation results and the output comparison results.

\subsection{Benchmark Evaluation}

\begin{table*}[tb]
    \centering\small
    \caption{Benchmark Evaluation Results}
    \label{tab:benchmark}
    \begin{tabular}{|c|c|c|c|c|c||c|}
        \hline
        Model        & chabsa (f1) & cma\_basics (acc) & cpa\_audit (acc)  & fp2 (acc)         & security\_sales\_1 (acc) & Overall   \\
        \hline
        Original     & $0.7381$    & $0.4737\pm0.0821$ & $0.1608\pm0.0184$ & $0.3389\pm0.0217$ & $0.4561\pm0.0666$        & $0.4335$  \\
        Ours (Tuned) & $0.7428$    & $0.5263\pm0.0821$ & $0.1633\pm0.0186$ & $0.3642\pm0.0221$ & $0.5614\pm0.0663$        & $0.4716$  \\
        \hline
        Diff         & $+0.0047$   & $+0.0526$         & $+0.0025$         & $+0.0253$         & $+0.1053$                & $+0.0381$ \\\hline
    \end{tabular}
\end{table*}

Table \ref{tab:benchmark} shows the benchmark evaluation results.
The tuned model achieved better performance than the original model in all tasks.
The overall score of the tuned model is $0.4716$, which is $+0.0381$ higher than the original model's score ($0.4335$).
The difference between the tuned model and the original model is not statistically significant, but it seems significant that the tuned model achieved better performance than the original model in all tasks.

\subsection{Outputs Comparison}
Only evaluating the benchmark results is not enough to understand the effectiveness of the tuning.
Therefore, we compared the outputs of the tuned model with those of the original model.
Here, we show some examples of the outputs comparison that the difference between the tuned model and the original model is huge or clear.
All the results are translated from Japanese to English.

\begin{brekableitembox}{Ex.1: Result starting with "The Bank of Japan"}
    {\bf Ours tuned (nekomata-14b-pfn-qfin):}
    The Bank of Japan introduced "Quantitative and Qualitative Monetary Easing" in April 2013 and has been promoting strong monetary easing to achieve the "price stability target" of 2 percent consumer price inflation. As a result, Japan's economy has been on a gradual recovery path, and the employment and income environment has been improving.
    Against this backdrop, corporate earnings are at record-high levels, and capital investment is on an increasing trend. In addition, personal consumption has remained steady, and housing investment is also picking up. Thus, the Japanese economy is judged to be recovering moderately.
    The Japanese economy is expected to expand moderately as the employment and income environment continues to improve steadily, and domestic demand is expected to remain firm, partly due to the effects of various policies. However, attention should be paid to the effects of uncertainties in overseas economies and fluctuations in financial markets. ...\\
    ---------------------------------------------------------------------\\
    {\bf Original (nekomata-14b):}
    The Bank of Japan has defined the objectives of its monetary policy as "price stability and financial system stability." Of these, with respect to "price stability," the Bank has set a target that the year-on-year rate of increase in the CPI (excluding fresh food) should be in the positive territory of 2\% or less, with a central target of around 1\%.
    On the other hand, with regard to "financial system stability," the Bank is supposed to work to stabilize the financial system through the resolution of bankruptcies of financial institutions and the development of systems to ensure financial system stability, as well as by exercising its function as a lender of last resort as a central bank.
    The Bank conducts its monetary policy operations in accordance with these missions. ...
\end{brekableitembox}

In Ex.1, the tuned model's output is slightly better than the original model's output because the central target of the price increase rate is correctly 2\% as the tuned model's output, but the original model's output says around 1\%.

\begin{brekableitembox}{Ex.2: Result starting with "Derivative instruments are"}
    {\bf Ours tuned (nekomata-14b-pfn-qfin):}
    Derivative instruments are transactions in financial instruments, such as stocks and bonds, that take advantage of price fluctuations to make a profit. For example, in a stock futures contract, you enter into a contract to buy or sell a stock at a specific price on a specific date in the future. Thus, you take a position to buy if you expect the stock price to rise and sell if you expect the stock price to fall. If the stock price moves as expected, you will earn a profit.
    There are many other types of derivative instruments. For example, an option is a transaction in which you buy or sell the right to buy or sell a specific quantity of a financial instrument at a particular price and within a specific time period. Swap transactions, on the other hand, take advantage of interest rate differences between different currencies and are used to hedge foreign exchange risk.
    Derivative instruments have the potential to generate significant profits for investors but also carry the risk of significant losses. ...\\
    ---------------------------------------------------------------------\\
    {\bf Original (nekomata-14b):}
    Derivative instruments are not guaranteed for the principal amount and may cause losses due to price fluctuations or interest rate fluctuations. In addition, the amount of loss may exceed the margin because the amount of the transaction is large compared to the amount of margin to be deposited by the customer for such transaction.
    Foreign exchange margin transactions (FX) may cause losses due to fluctuations in the price of a currency or the value of a financial index, and the amount of such losses may exceed the amount of margins deposited by a customer for such transactions. Furthermore, the amount of loss may exceed the margin because the transaction amount is large compared to the amount of margin to be deposited by the customer for such transaction.
\end{brekableitembox}

In Ex.2, the tuned model's output is much better than the original model's because the tuned model's output correctly explains the derivative instruments, while the original model's output does not.
Moreover, the tuned model's output is more informative than the original model's output, and its output's length is longer than the original model's output (The full output of the tuned model is not shown in the box, but the output of the original model is fully shown in the box).

\begin{brekableitembox}{Ex.3: Result starting with "A knockout option is"}
    {\bf Ours tuned (nekomata-14b-pfn-qfin):}
    A knockout option is a type of binary option, a trade in which the trader predicts whether the decision price will be above or below the strike price at the expiration time. FX account applications will not be accepted between 6:55 a.m. Thursday, January 3, 2019, and 6:55 a.m. Friday, January 4, 2019, due to system maintenance.
    Select Month August 2019 (170) July 2019 (200) June 2019 (200) May 2019 (200) April 2019 (200) March 2019 (140) February 2019 (200) January 2019 (160) December 2018 (200) November 2018 (160)\\
    ---------------------------------------------------------------------\\
    {\bf Original (nekomata-14b):}
    A knockout option is a type of binary option that predicts whether the price of the underlying asset will be above (high) or below (low) the target rate at the decision time. January 3, 2019 Virtual currency exchanges and sellers offer a variety of altcoins in addition to Bitcoin: Ethereum is one of the most popular altcoins. Like Bitcoin, Ethereum is a virtual currency that uses blockchain technology but differs from Bitcoin in some ways. In this article, we will discuss the features of Ethereum, how it works, and how to purchase it.
    2018/07/16 2019/04/15 2019/04/15 2019/04/15 2019/04/15 2019/04/15 2019/04/15 2019/04/15 2019/04/15 2019/04/15 2019/04/15 2019/04/15 2019/04/15 2019/04/15 2019/04/15 2019/04/15 2019/04/15 2019/04/15 2019/04/15 2019/04/15 2019/04/15 2019/04/15 2019/04/15 2019/04/15 2019/04/15 2019/04/15 2019/04/15 2019/04/15
\end{brekableitembox}

Another example is Ex.3. In this example, both the tuned model's output and the original model's output are not correct (a knockout option is not a type of binary option).
Moreover, both outputs contain irrelevant information.
So, even the tuned model cannot answer correctly in some cases for financial domain-specific questions.

\begin{brekableitembox}{Ex.4: Result starting with "VIX is"}
    {\bf Ours tuned (nekomata-14b-pfn-qfin):}
    VIX is a "fear index" that reflects investor sentiment toward the future of the US stock market. It usually hovers between 10 and 20, with higher values indicating greater caution about the future.
    The VIX is calculated from options trading linked to the S\&P 500. The VIX is calculated by the Chicago Board Options Exchange (CBOE) and is commonly referred to as the "fear index"; a VIX above 20 is considered to indicate an increase in market volatility, while a VIX above 30 is viewed as a danger zone.
    The US stock market continues to be unstable against the backdrop of concerns over the global economic slowdown and low oil prices due to the spread of the new coronavirus. After reaching an all-time high on February 12, the Dow Jones Industrial Average fell to a record low of \$2,997 on February 24. The market continued to fluctuate wildly and fell below the \$20,000 level again on March 6. Against this backdrop, the VIX temporarily rose to 50.30 on March 11, the highest level in about three years. The VIX can also fluctuate significantly depending on US monetary policy and political developments. For example, ...\\
    ---------------------------------------------------------------------\\
    {\bf Original (nekomata-14b):}
    VIX is an index that indicates the volatility of the US stock market, also known as the "fear index." The higher the value, the more anxious investors become, and when the value exceeds 20, it is considered a level comparable to a financial crisis.
    In the US stock market on August 18, the Dow Jones Industrial Average closed at \$24,608, down \$545 from the previous day. The drop temporarily exceeded \$700. Selling swelled mainly in China-related stocks due to concerns over the trade friction between the US and China.
\end{brekableitembox}

Ex.4 is an interesting example.
In this example, both outputs seem correct but include different information from just explaining VIX.
However, the explanation for VIX by the tuned model is more informative than the original model's explanation.

\begin{brekableitembox}{Ex.5: Result starting with "A weaker yen is"}
    {\bf Ours tuned (nekomata-14b-pfn-qfin):}
    A weaker yen is a positive factor for the Japanese economy. However, a weaker yen raises the prices of imported goods, pushing up consumer prices. In addition, for exporters, a weaker yen increases the price competitiveness of products sold overseas, thereby increasing their earnings. On the other hand, for importers, a weaker yen increases their purchasing costs, resulting in lower earnings.
    Thus, the impact of the yen's depreciation cannot be said to be either good or bad since profits and losses are divided between exporters and importers. In general, however, a weaker yen is considered favorable for exporters and unfavorable for importers.\\
    \\
    Comment 2\\
    While a weak yen is beneficial for export- and inbound-related industries, ...\\
    ---------------------------------------------------------------------\\
    {\bf Original (nekomata-14b):}
    A weaker yen is positive for the performance of exporting firms. It may also make it easier for foreign investors to buy Japanese stocks, leading to higher stock prices.
    On the other hand, there is a fear that the rate of increase in the consumer price index (CPI) could accelerate through higher import prices. The BOJ forecasts the CPI for FY 2019 at 0.8\% y/y, but if the yen continues to weaken, this forecast may not be achieved.
\end{brekableitembox}

Also, in Ex.5, the tuned model's output is more informative than the original model's output.
Moreover, the tuned model's output only focuses on the impact of the weaker yen, while the original model's output also includes information on stock prices and the CPI.
However, the format of the tuned model's output seems weird because it contains "Comment 2," and the following sentence is unrelated to the previous sentence.

According to those examples, the tuned model's outputs tend to be better than the original model's outputs in terms of the quality and length of the answers.
However, the example we showed here is very limited, and many other comparison results do not have a significant difference between the tuned model and the original model.

\section{Discussion}
First, the benchmark evaluation results show that the tuned model performed better than the original model in all tasks.
The dataset size is not so large, but these results show that our tuning was archived to add the knowledge of the financial domain to the model.

This tendency is also shown in the outputs comparison.
For example, the output results of Ex.2 and 4 also indicated that the tuned model's outputs are better than the original model's outputs in terms of correctness and informativeness.

However, the datasets used for the tuning are not so large but specially focused on the financial domain, which could be the main reason that the tuned model achieved better performance than the original model.
Like the domain-specified BERTs, the domain-specific tuning could be effective for the LLMs.

On the other hand, the outputs comparison results also showed that the tuned model still had issues answering some questions correctly.
For example, the output results of Ex.3 indicated that the tuned model could not answer correctly in some cases for financial domain-specific questions.
Moreover, the benchmark is also not a full score, so the tuned model is not perfect yet in terms of financial knowledge.
In addition to the knowledge issue, LLM-specific issues, such as hallucination, still exist.

Some possible future works exist to address those issues.
To address the knowledge issue, the dataset for finance should be more diverse and larger.
Moreover, instruction tuning \cite{wei2021finetuned} could be another future work.
Currently, our tuned model only supports generating continual text, but instruction tuning could be effective for question-answering tasks.
The instruction tuning could also ease the hallucination issue.
Therefore, the instruction datasets and tuning focusing on the financial domain could be vital for future research.

According to our results and discussion, domain-specific tuning is also effective for LLMs, but it is not clear that the tuning is effective even for LLMs with huge parameters, such as 100-billion-class-parameter models.
GPT-4 series is one of the 100 billion-class-parameter models, and its benchmark score is far better than our tuned model's.
Therefore, the effectiveness of the domain-specific tuning for the 100-billions-class-parameters models is still unclear.
Therefore, future work should also include the evaluation of the domain-specific tuning for the 100-billions-class-parameters models.

\section{Conclusion}
This study aims to construct a Japanese financial-specific LLM.
For the tuning, we employed the continual pre-training method.
Before tuning, we constructed Japanese financial-focused datasets for the continual pre-training containing about 8.1 million documents and 370 million tokens.
As a base model, we employed rinna/nekomata-14b, publicly available on Hugging Face, and achieved state-of-the-art performance on Japanese financial benchmarks among the 10-billions-class-parameters models.
Then, we performed continual pre-training using the datasets and the base model.
As evaluations, we employed the Japanese financial benchmarks and the outputs comparison.
The results reveal that the tuned model performed better than the original model in all benchmarks.
Moreover, the outputs comparison results also showed that the tuned model's outputs tend to be better than the original model's outputs in terms of the quality and length of the answers.
However, the tuned model still has issues to answer correctly for some questions.
According to these results, the domain-specific tuning is still effective for the LLMs.
Finally, the scope for future research includes instruction tuning, additional datasets covering broader financial knowledge, and the evaluation of domain-specific tuning for the 100-billions-class-parameter models.

\bibliographystyle{IEEEtran}
\bibliography{cite}

\begin{thebibliography}{10}
\providecommand{\url}[1]{#1}
\csname url@samestyle\endcsname
\providecommand{\newblock}{\relax}
\providecommand{\bibinfo}[2]{#2}
\providecommand{\BIBentrySTDinterwordspacing}{\spaceskip=0pt\relax}
\providecommand{\BIBentryALTinterwordstretchfactor}{4}
\providecommand{\BIBentryALTinterwordspacing}{\spaceskip=\fontdimen2\font plus
\BIBentryALTinterwordstretchfactor\fontdimen3\font minus \fontdimen4\font\relax}
\providecommand{\BIBforeignlanguage}[2]{{%
\expandafter\ifx\csname l@#1\endcsname\relax
\typeout{** WARNING: IEEEtran.bst: No hyphenation pattern has been}%
\typeout{** loaded for the language `#1'. Using the pattern for}%
\typeout{** the default language instead.}%
\else
\language=\csname l@#1\endcsname
\fi
#2}}
\providecommand{\BIBdecl}{\relax}
\BIBdecl

\bibitem{chatgpt}
OpenAI, ``{ChatGPT},'' 2023, \url{https://openai.com/blog/chatgpt/}.

\bibitem{GPT4}
\BIBentryALTinterwordspacing
OpenAI, ``{GPT-4 Technical Report},'' 2023. [Online]. Available: \url{\url{https://arxiv.org/abs/2303.08774}}
\BIBentrySTDinterwordspacing

\bibitem{Vaswani2017}
A.~Vaswani, N.~Shazeer, N.~Parmar, J.~Uszkoreit, L.~Jones, A.~N. Gomez, {\L}.~Kaiser, and I.~Polosukhin, ``{Attention Is All You Need},'' in \emph{Advances in Neural Information Processing Systems}, vol.~30, 2017, pp. 5999--6009.

\bibitem{Devlin2018}
J.~Devlin, M.-W. Chang, K.~Lee, and K.~Toutanova, ``{BERT: Pre-training of Deep Bidirectional Transformers for Language Understanding},'' in \emph{Proceedings of the 2019 Conference of the North {A}merican Chapter of the Association for Computational Linguistics}.\hskip 1em plus 0.5em minus 0.4em\relax Association for Computational Linguistics, 2019, pp. 4171--4186.

\bibitem{GPT-1}
A.~Radford, K.~Narasimhan, T.~Salimans, and I.~Sutskever, ``{Improving Language Understanding by Generative Pre-Training},'' 2018, \url{https://cdn.openai.com/research-covers/language-unsupervised/language_understanding_paper.pdf}.

\bibitem{GPT-2}
A.~Radford, J.~Wu, R.~Child, D.~Luan, D.~Amodei, and I.~Sutskever, ``{Language Models are Unsupervised Multitask Learners},'' 2019, \url{https://cdn.openai.com/better-language-models/language_models_are_unsupervised_multitask_learners.pdf}.

\bibitem{GPT-3}
T.~Brown, B.~Mann \emph{et~al.}, ``{Language Models are Few-Shot Learners},'' \emph{Advances in Neural Information Processing Systems}, vol.~33, pp. 1877--1901, 2020.

\bibitem{bard}
Google, ``Bard,'' 2023, \url{https://bard.google.com/}.

\bibitem{touvron2023llama}
H.~Touvron, T.~Lavril, G.~Izacard, X.~Martinet, M.-A. Lachaux, T.~Lacroix, B.~Rozi{\`e}re, N.~Goyal, E.~Hambro, F.~Azhar \emph{et~al.}, ``{LLaMA: Open and Efficient Foundation Language Models},'' \emph{arXiv}, 2023, \url{https://arxiv.org/abs/2302.13971}.

\bibitem{Touvron2023}
H.~Touvron, L.~Martin \emph{et~al.}, ``{Llama 2: Open Foundation and Fine-Tuned Chat Models},'' \emph{arXiv}, 2023, \url{https://arxiv.org/abs/2307.09288v2}.

\bibitem{dolly}
Databricks, ``Dolly,'' 2023, \url{https://github.com/databrickslabs/dolly}.

\bibitem{scao2022bloom}
T.~L. Scao, A.~Fan, C.~Akiki, E.~Pavlick, S.~Ili{\'c}, D.~Hesslow, R.~Castagn{\'e}, A.~S. Luccioni, F.~Yvon, M.~Gall{\'e} \emph{et~al.}, ``{BLOOM: A 176B-Parameter Open-Access Multilingual Language Model},'' \emph{arXiv}, 2022, \url{https://arxiv.org/abs/2211.05100}.

\bibitem{vicuna}
Vicuna, ``{Vicuna: An Open-Source Chatbot Impressing GPT-4 with 90\%* ChatGPT Quality},'' 2023, \url{https://vicuna.lmsys.org/}.

\bibitem{Chowdhery2022}
A.~Chowdhery, S.~Narang \emph{et~al.}, ``{PaLM: Scaling Language Modeling with Pathways},'' \emph{arXiv}, 2022, \url{https://arxiv.org/abs/2204.02311v5}.

\bibitem{Anil2023}
R.~Anil, A.~M. Dai \emph{et~al.}, ``{PaLM 2 Technical Report},'' \emph{arXiv}, 2023, \url{https://arxiv.org/abs/2305.10403v3}.

\bibitem{gemini}
G.~Team, ``Gemini: a family of highly capable multimodal models,'' \emph{arXiv preprint arXiv:2312.11805}, 2023.

\bibitem{Suzuku2023-ipm}
M.~SUZUKI, H.~SAKAJI, M.~HIRANO, and K.~IZUMI, ``{Constructing and Analyzing Domain-Specific Language Model for Financial Text Mining},'' p. e103194, 2023.

\bibitem{Hirano2023-nbis}
M.~HIRANO, M.~SUZUKI, and H.~SAKAJI, ``{llm-japanese-dataset v0: Construction of Japanese Chat Dataset for Large Language Models and its Methodology},'' in \emph{The 26th International Conference on Network-Based Information Systems}, 2023, pp. 442--454.

\bibitem{sukeda2023jmedlora}
I.~Sukeda, M.~Suzuki, H.~Sakaji, and S.~Kodera, ``{JMedLoRA: Medical Domain Adaptation on Japanese Large Language Models using Instruction-tuning},'' \emph{arXiv}, 2023, \url{https://arxiv.org/abs/2310.10083}.

\bibitem{beltagy2019scibert}
I.~Beltagy, K.~Lo, and A.~Cohan, ``Scibert: A pretrained language model for scientific text,'' in \emph{Proceedings of the 2019 Conference on Empirical Methods in Natural Language Processing and the 9th International Joint Conference on Natural Language Processing (EMNLP-IJCNLP)}, 2019, pp. 3615--3620.

\bibitem{rasmy2021med}
L.~Rasmy, Y.~Xiang, Z.~Xie, C.~Tao, and D.~Zhi, ``Med-bert: pretrained contextualized embeddings on large-scale structured electronic health records for disease prediction,'' \emph{NPJ digital medicine}, vol.~4, no.~1, p.~86, 2021.

\bibitem{Suzuki2022-sigfin28}
\BIBentryALTinterwordspacing
M.~SUZUKI, H.~SAKAJI, M.~HIRANO, and K.~IZUMI, ``{Construction and Validation of a Pre-Training and Additional Pre-Training Financial Language Model [in Japanese]},'' in \emph{The 28th meeting of Special Interest Group on Financial Informatics of Japanese Society for Artificial Intelligence}, 2022, pp. 132--137. [Online]. Available: \url{https://sigfin.org/?028-24}
\BIBentrySTDinterwordspacing

\bibitem{howard2018ulmfit}
J.~Howard and S.~Ruder, ``Universal language model fine-tuning for text classification,'' in \emph{Proceedings of the 56th Annual Meeting of the Association for Computational Linguistics (Volume 1: Long Papers)}.\hskip 1em plus 0.5em minus 0.4em\relax Association for Computational Linguistics, 2018, pp. 328--339.

\bibitem{gururangan2020don}
S.~Gururangan, A.~Marasovi{\'c}, S.~Swayamdipta, K.~Lo, I.~Beltagy, D.~Downey, and N.~A. Smith, ``Don't stop pretraining: Adapt language models to domains and tasks,'' \emph{arXiv preprint arXiv:2004.10964}, 2020.

\bibitem{Wu2023}
S.~Wu, O.~Irsoy, S.~Lu, V.~Dabravolski, M.~Dredze, S.~Gehrmann, P.~Kambadur, D.~Rosenberg, and G.~Mann, ``{BloombergGPT: A Large Language Model for Finance},'' \emph{arXiv}, 2023, \url{https://arxiv.org/abs/2303.17564v2}.

\bibitem{Fin-LLAMA}
P.~B. William~Todt, Ramtin~Babaei, ``{Fin-LLAMA: Efficient Finetuning of Quantized LLMs for Finance},'' 2023, \url{https://github.com/Bavest/fin-llama}.

\bibitem{yang2023fingpt}
H.~Yang, X.-Y. Liu, and C.~D. Wang, ``{FinGPT: Open-Source Financial Large Language Models},'' \emph{arXiv}, 2023, \url{https://arxiv.org/abs/2306.06031}.

\bibitem{zhang2023instruct}
B.~Zhang, H.~Yang, and X.-Y. Liu, ``{Instruct-FinGPT: Financial Sentiment Analysis by Instruction Tuning of General-Purpose Large Language Models},'' \emph{arXiv}, 2023, \url{https://arxiv.org/abs/2306.12659}.

\bibitem{Hirano2023-finnlpkdf}
M.~Hirano, ``{Construction of a Japanese Financial Benchmark for Large Language Models},'' in \emph{Joint Workshop of the 7th Financial Technology and Natural Language Processing (FinNLP), the 5th Knowledge Discovery from Unstructured Data in Financial Services (KDF), and The 4th Workshop on Economics and Natural Language Processing (ECONLP)}, 2024.

\bibitem{Mikolov2013a}
T.~Mikolov, K.~Chen, G.~Corrado, and J.~Dean, ``{Distributed Representations of Words and Phrases and their Compositionality},'' in \emph{Advances in Neural Information Processing Systems (NeurIPS)}, vol.~26, 2013, pp. 3111--3119.

\bibitem{Hirano2019-information}
M.~HIRANO, H.~SAKAJI, S.~KIMURA, K.~IZUMI, H.~MATSUSHIMA, S.~NAGAO, and A.~KATO, ``{Related Stocks Selection with Data Collaboration Using Text Mining},'' p. e102, 2019.

\bibitem{peters2018elmo}
M.~E. Peters, M.~Neumann, M.~Iyyer, M.~Gardner, C.~Clark, K.~Lee, and L.~Zettlemoyer, ``Deep contextualized word representations,'' in \emph{Proceedings of the 2018 Conference of the North {A}merican Chapter of the Association for Computational Linguistics: Human Language Technologies}, vol.~1.\hskip 1em plus 0.5em minus 0.4em\relax Association for Computational Linguistics, 2018, pp. 2227--2237.

\bibitem{schuster1997bilstm}
M.~Schuster and K.~Paliwal, ``Bidirectional recurrent neural networks,'' \emph{IEEE Transactions on Signal Processing}, vol.~45, no.~11, pp. 2673--2681, 1997.

\bibitem{suzuki2023base}
M.~Suzuki, M.~Hirano, and H.~Sakaji, ``{From Base to Conversational: Japanese Instruction Dataset and Tuning Large Language Models},'' \emph{arXiv}, 2023, \url{https://arxiv.org/abs/2309.03412}.

\bibitem{jlm_eval}
StabilityAI, ``{JP Language Model Evaluation Harness},'' 2023, \url{https://github.com/Stability-AI/lm-evaluation-harness/tree/jp-stable}.

\bibitem{llm-jp-eval}
\BIBentryALTinterwordspacing
LLM-jp, ``{llm-jp-eval},'' 2024. [Online]. Available: \url{https://github.com/llm-jp/llm-jp-eval}
\BIBentrySTDinterwordspacing

\bibitem{hu2021lora}
E.~J. Hu, Y.~Shen, P.~Wallis, Z.~Allen-Zhu, Y.~Li, S.~Wang, L.~Wang, and W.~Chen, ``Lora: Low-rank adaptation of large language models,'' \emph{arXiv preprint arXiv:2106.09685}, 2021.

\bibitem{wei2021finetuned}
J.~Wei, M.~Bosma, V.~Y. Zhao, K.~Guu, A.~W. Yu, B.~Lester, N.~Du, A.~M. Dai, and Q.~V. Le, ``Finetuned language models are zero-shot learners,'' \emph{arXiv preprint arXiv:2109.01652}, 2021.

\bibitem{christiano2017deep}
P.~F. Christiano, J.~Leike, T.~Brown, M.~Martic, S.~Legg, and D.~Amodei, ``Deep reinforcement learning from human preferences,'' \emph{Advances in neural information processing systems}, vol.~30, 2017.

\bibitem{rafailov2024direct}
R.~Rafailov, A.~Sharma, E.~Mitchell, C.~D. Manning, S.~Ermon, and C.~Finn, ``Direct preference optimization: Your language model is secretly a reward model,'' \emph{Advances in Neural Information Processing Systems}, vol.~36, 2024.

\bibitem{zhou2024lima}
C.~Zhou, P.~Liu, P.~Xu, S.~Iyer, J.~Sun, Y.~Mao, X.~Ma, A.~Efrat, P.~Yu, L.~Yu \emph{et~al.}, ``Lima: Less is more for alignment,'' \emph{Advances in Neural Information Processing Systems}, vol.~36, 2024.

\bibitem{accelerate}
S.~Gugger, L.~Debut, T.~Wolf, P.~Schmid, Z.~Mueller, S.~Mangrulkar, M.~Sun, and B.~Bossan, ``Accelerate: Training and inference at scale made simple, efficient and adaptable.'' \url{https://github.com/huggingface/accelerate}, 2022.

\bibitem{rasley2020deepspeed}
J.~Rasley, S.~Rajbhandari, O.~Ruwase, and Y.~He, ``Deepspeed: System optimizations enable training deep learning models with over 100 billion parameters,'' in \emph{Proceedings of the 26th ACM SIGKDD International Conference on Knowledge Discovery \& Data Mining}, 2020, pp. 3505--3506.

\bibitem{maini2024rephrasing}
P.~Maini, S.~Seto, H.~Bai, D.~Grangier, Y.~Zhang, and N.~Jaitly, ``Rephrasing the web: A recipe for compute and data-efficient language modeling,'' \emph{arXiv preprint arXiv:2401.16380}, 2024.

\bibitem{Masuda2023}
T.~Masuda, K.~Nakagawa, and T.~Hoshino, ``Can chatgpt pass the jcpa exam?: Challenge for the short-answer method test on auditing,'' in \emph{The 31st meeting of Special Interest Group on Financial Informatics of Japanese Society for Artificial Intelligence}, 2023, pp. 81--88.

\end{thebibliography}

\end{document}